# Black-Box Complexities of Combinatorial Problems[*]

Benjamin Doerr[1], Timo Kötzing[1], Johannes Lengler[2], Carola Winzen[1]

[1]Max-Planck-Institut für Informatik, Saarbrücken, Germany
[2]ETH Zürich, Zürich, Switzerland

### Abstract

Black-box complexity is a complexity theoretic measure for how difficult a problem is to be optimized by a general purpose optimization algorithm. It is thus one of the few means trying to understand which problems are tractable for genetic algorithms and other randomized search heuristics.

Most previous work on black-box complexity is on artificial test functions. In this paper, we move a step forward and give a detailed analysis for the two combinatorial problems minimum spanning tree and single-source shortest paths. Besides giving interesting bounds for their black-box complexities, our work reveals that the choice of how to model the optimization problem is non-trivial here. This in particular comes true where the search space does not consist of bit strings and where a reasonable definition of unbiasedness has to be agreed on.

## 1 Introduction

Black-box complexity is a notion trying to capture how difficult a problem is to be solved via problem-independent, possibly randomized, search heuristics. Roughly speaking, the black-box complexity of a problem (a class of functions to be optimized) is the expected number of function evaluations needed to find the optimum of an unknown member of the class (Droste, Jansen, and Wegener [DJW06]).

This *unrestricted* black-box model sometimes gives unrealistically small complexity values (as compared with run times exhibited by standard randomized search heuristics (RSH)). A way to overcome this is to restrict the class of randomized algorithms regarded (of course, in a way that classic RSH are still included). To this aim, Lehre and Witt [LW10] suggested an *unbiased black-box* model, in which algorithms are only allowed to generate new solutions from existing ones, and only via so-called unbiased variation operators. Doerr and Winzen [DW11] regard the restriction that the algorithm has no access to the absolute objective values of solutions, but only to the ranking implied by their fitnesses. This leads to an (unrestricted or unbiased) *ranking-based black-box* model.

A number of deep and sometimes unexpected results exist for the different notions, most of them, however, only regarding artificial test problems like OneMax, LeadingOnes, or jump functions. The focus of this paper is to start an in-depth analysis of black-box complexities for combinatorial problems. As we will see in this paper, a number of additional modeling issues have to be regarded here. We start our analysis with the minimum spanning tree (MST) problem, because here it is generally agreed on that a bit-string representation is most natural. This allows to use the definition of unbiasedness as in [LW10]. When talking about ranking-based black-box complexity, the two-criteria fitness (total weight, number of connected components) needs attention, but the only reasonable model is to treat the two criteria separately, i.e., to assume comparability in both criteria.

---

[*]This is the full version of the one presented at the Genetic and Evolutionary Computation Conference (GECCO 2011) [DKLW11]. A journal version is currently in preparation.

We then proceed to the single-source shortest path (SSSP) problem, where current-best evolutionary approaches use representations different from bit-strings. Here it is not clear a priori what unbiasedness shall mean. Transforming the definition from Lehre and Witt [LW10] in a straight-forward way leads to not very useful results. Taking the problem semantics into account, we find a reasonable definition for unbiasedness and prove meaningful black-box complexities.

**Spanning Trees.** In one of the earliest theoretical works on evolutionary algorithms for combinatorial optimization problems, Neumann and Wegener [NW04, NW07] analyze the (expected) optimization time of the (1+1) evolutionary algorithm (EA) for the MST problem. They prove that the expected time to find one is $O(m^2 \log(nw_{\max}))$, where $n$ is the number of vertices, $m$ the number of edges and $w_{\max}$ is the maximum of the positive and integral edge weights. It is a major open problem whether the dependence on the maximum edge weight is necessary.

The same bound is proven for a randomized local search (RLS) variant doing one-bit and two-bit flips each with probability 1/2. This can be easily improved to $O(m^2 \log n)$ by noting that the optimization behavior remains exactly the same if we replace the existing edge weights by the numbers from 1 to $m$ (keeping the relative order of the edge weights unchanged) [RS09].

Since the MST problem has a natural representation via bit strings, for this combinatorial problem we can easily use the four black-box complexity notions. Our results can be found in Table 1. They imply that the unbiased black-box complexity is asymptotically different for the unary case and for arities $\geq 3$.

In a nutshell, the results show that, on the one hand, simple algorithms based on unary operators, such as EAs and RLS can get run times very close to the theoretically optimal; on the other hand, they show how operators of higher arity can further improve on the run time (see also [DHK08, DT09, DJK$^+$10] for higher-arity operators for combinatorial optimization problems).

**Shortest Paths.** In another one of the earliest theoretical works on evolutionary algorithms for combinatorial optimization problems, Scharnow, Tinnefeld, and Wegener [STW02, STW04] analyze how a (1+1) EA solves the SSSP problem.

Since in the SSSP problem a shortest path between the source and any other vertex is sought for, a bit-string representation for solution candidates seems not very natural. Therefore, most works resort to trees or slightly more general structures as representations. To ease the comparison with most existing works on the SSSP problem, in this work we shall only work with the vertex-based representation employed in [STW04], which, roughly speaking, for each vertex stores its predecessor on the path from the source to it. We note that superior run times were recently proven for an edge-based approach [DJ10].

In addition, also the choice of the fitness function is subtle. In [STW04], a *multi-criteria fitness* was suggested. For each vertex, the objective function returns the distance from the source in the current solution (infinity, if the vertex is not connected to the source). An offspring is only accepted if, in each of these $n-1$ criteria, it is not worse than the parent. For the natural (1+1) EA building on this framework, they prove an expected optimization time of $O(n^3)$. This was improved to a bound of $O(n^2 \max\{\ell, \log(n)\})$, where $\ell$ is the smallest height of a shortest path tree [DHK07].

|              | (rb) unrestr. | *-ary unb.  | unary unb.           | rb unary unb.        | (rb) binary unb. | (rb) 3-ary unb. |
|--------------|---------------|-------------|----------------------|----------------------|------------------|-----------------|
| upper bound  | $2m+1$        | $O(m)$      | $O(mn \log n)^\dagger$ | $O(mn(\log n)^2)$   | $O(m \log n)$    | $O(m)$          |
| lower bound  | $(1-o(1))n$   | $\Omega(n)$ | $\Omega(m \log n)$   | $\Omega(m \log n)$   | $\Omega(m/\log n)$ | $\Omega(m/\log n)$ |

Table 1: Upper and lower bounds for the black-box complexity of MST in the different models. Abbreviations: unrestr.= unrestricted, rb = ranking-based, unb. = unbiased.
$^\dagger$ $O(mn \log(m/n))$ if all edge weights are distinct.

When analyzing the black-box complexity of this formulation of the SSSP problem, we first note that both unbiased and ranking-based complexities make little sense. Since the multi-objective fitness explicitly distinguishes the vertices, treating vertices equally here (as done by unbiased operators) or making individual distances incomparable (as done by component-wise ranking) is ill-natured.

Hence for the multi-criteria fitness, we shall only regard the unrestricted black-box complexity. Interestingly, this problem is also among the few combinatorial problems for which black-box complexity results exist. Droste, Jansen, (Tinnefeld,) and Wegener [DJTW03, DJW06] showed that the unrestricted black-box complexity of the SSSP in the multi-criteria formulation is at least $n/2$ and at most $2n - 3$.[1] We first improve these bounds to exactly $n - 1$ for both the upper and the lower bound.[2] Surprisingly, if we may assume that the input graph is a complete graph, we obtain a black-box complexity of at most $n/2 + O(1)$, see Table 3 in Section 4.1. That is, the SSSP problem becomes easier (in the black-box complexity sense) if we transform an arbitrary instance to one on a complete graph (but adding expensive dummy edges).

The natural *single-criterion* formulation of the SSSP problem takes as objective simply the sum of the distances of all vertices to the source in the current solution. This approach was dismissed in [STW04] for the reason that then all solutions with at least one vertex not connected to the source form a huge plateau of equal fitness.

In [BBD$^+$09], it was observed that this (artificial) problem dissolves if each unconnected vertex only contributes a large value (e.g., larger than the sum of all edge weights) to the objective value. This is the common way to implement the $\infty$-value in most algorithms. In this setting, also the single-criterion EA is efficient and finds the optimum, on average, in $O(n^3 \log(nw_{\max}))$ iterations.

For the single-criterion version of the SSSP problem, there is no reason to not regard unbiased black-box complexities. However, we shall see that finding a good notion for unbiasedness is a crucial point here. From the representation point of view, since individuals are nothing more than certain mappings from the vertex set into itself, unbiasedness in the sense of Lehre and Witt would mean that we treat all possible images of each vertex symmetrically. From the problem view-point, unbiased should mean that we treat all vertices (apart from the source) equal. This is a substantial difference, as we shall discuss in detail in Section 4.2. Both approaches lead to different black-box complexities, cf. Table 2.

---

[1] The only other result published on the black-box complexity of a combinatorial problem is the proof that the $NP$-complete MaxClique problem has a polynomial black-box complexity, see again [DJW06].

[2] Note that the upper bound in [DJW06] still holds in a more restricted setting, cf. Section 4.1.

|  | unrestr. | rb unrestr. | unary struct | binary struct | 3-ary struct | unary redir |
|---|---|---|---|---|---|---|
| upper bound | $n(n-1)/2$ | $(n-1)^2$ | $O(n^3 \log n)$ | $O(n^2 \log n)$ | $O(n^2)$ | $O(n^3)$ |

Table 2: Upper bounds for the black-box complexity of SSSP with single-criteria fitness function in the different models. Abbreviations: unrestr.= unrestricted, rb = ranking-based, struct = structure preserving unbiased, redir = redirecting unbiased.

## 2 The Four Black-Box Models

In this section we present two black-box models, the *unrestricted* black-box model by Droste et al. [DJW06] and the *unbiased* model by Lehre and Witt [LW10]. Furthermore, we shall introduce the unrestricted and the unbiased *ranking-based* black-box models in Section 2. We give only a short presentation of the models. For a more detailed exposure confer [DJW06, DW11, LW10]. The description below follows closely the one in [DKW11].

Throughout this work, we use the following notations. We denote the positive integers by $\mathbb{N}$ and the positive reals by $\mathbb{R}^+$. For any $k \in \mathbb{N}$, we abbreviate $[k] := \{1, \ldots, k\}$. Analogously, we define $[0..k] := [k] \cup \{0\}$. For $x = x_1 \cdots x_n \in \{0,1\}^n$ we denote by $\bar{x}$ the bitwise complement of $x$ (i.e., for all $i \in [n]$ we have $\bar{x}_i = 1 - x_i$). The bitwise exclusive OR is denoted by $\oplus$. For any set $S$ we denote by $2^S$ the power set of $S$, i.e., the set of all subsets of $S$. By $S_n$ we denote the set of all permutations of $[n]$. Lastly, with $\log_a$ we denote the logarithm to base $a$, and with log we denote the natural logarithm to base $e := \exp(1)$.

**The Unrestricted and Unbiased Black-Box Models.** We are interested in measuring the complexity of a problem's *optimizability by randomized search heuristics*. Black-box complexity follows the usual approach to take as a measure the *performance of the best algorithm* out of some class of algorithms. As our main interest is in the performance of RSH, we restrict our attention to the class of algorithms which obtain information about the problem to be solved only by learning the objective value of possible solutions. The objective function is given by an oracle, or as a *black-box*. Using this oracle, the algorithm may query the objective value of all possible solutions, but any such query does only return this solution's objective value and no other information about the objective function.

Naturally, we allow that the algorithms are adaptive and that they use random decisions. However, the only type of action the algorithm may perform is, based on the objective values learned so far, deciding on a probability distribution on the search space $\mathcal{S}$, sampling from it a solution ("search point") $x \in \mathcal{S}$, and querying its objective value ("fitness") from the oracle. This leads to the black-box model by Droste et al. [DJW06] which contains all algorithms following the scheme of an *unrestricted black-box algorithm*, cf. Algorithm 1.

In typical applications of RSH, the evaluation of the fitness of a search point is more costly than the generation of a new one. Thus, we take as performance measure of a black-box algorithm the number of queries to the oracle until the algorithm first queries an optimal solution. Since we mainly talk about randomized algorithms, we regard the *expected* number of such queries and call this value the *run time* of the black-box algorithm.

For a class $\mathcal{F}$ of functions, the *complexity* of an algorithm $A$ for $\mathcal{F}$ is the worst-case run time, i.e., the maximum run time of $A$ on a function $f \in \mathcal{F}$. The complexity of $\mathcal{F}$ with respect to a class $\mathcal{A}$ of algorithms is the minimum ("best") complexity among all $A \in \mathcal{A}$ for $\mathcal{F}$. Hence, the *unrestricted black-box complexity* of $\mathcal{F}$ is the complexity of $\mathcal{F}$ with respect to the class of all unrestricted black-box algorithms.

---
**Algorithm 1:** Scheme of an Unrestricted Black-Box Algorithm
---
1 **Initialization:** Sample $x^{(0)}$ according to some probability distribution $p^{(0)}$ on $\mathcal{S}$. Query $f(x^{(0)})$.
2 **Optimization: for** $t = 1, 2, 3, \ldots$ **until** *termination condition met* **do**
3     Depending on $((x^{(0)}, f(x^{(0)})), \ldots, (x^{(t-1)}, f(x^{(t-1)})))$ choose a probability distribution $p^{(t)}$ on $\mathcal{S}$.
4     Sample $x^{(t)}$ according to $p^{(t)}$, and query $f(x^{(t)})$.

---

Already the authors of [DJW06] noted that the unrestricted black-box is very powerful. As an example, consider a single objective function $f$. Clearly, the unrestricted black-box complexity of $\{f\}$ is 1 — the algorithm which queries an optimal solution of $f$ as first action shows this bound.

This motivated Lehre and Witt [LW10] to introduce a more restrictive black-box model, where algorithms may generate new solution candidates only from random or previously generated search points and only by using *unbiased* variation operators. Note already here that Lehre and Witt formulated their model only for the hypercube $\{0,1\}^d$ as search space. In Section 4, we propose two ways to carry over the notion to a different setting. Next is a brief presentation of the model by Lehre and Witt.

**Definition 1.** *Let $k \in \mathbb{N}$. A k-ary* unbiased distribution *is a family of probability distributions* $\left(D(\cdot \mid y^{(1)}, \ldots, y^{(k)})\right)_{y^{(1)}, \ldots, y^{(k)} \in \{0,1\}^d}$ *over $\{0,1\}^d$ such that for all inputs $y^{(1)}, \ldots, y^{(k)} \in \{0,1\}^d$ the following two conditions hold.*

$$(i)\, \forall x, z \in \{0,1\}^d: \quad D(x \mid y^{(1)}, \ldots, y^{(k)}) = D(x \oplus z \mid y^{(1)} \oplus z, \ldots, y^{(k)} \oplus z);$$
$$(ii)\, \forall x \in \{0,1\}^d\, \forall \sigma \in S_n: \quad D(x \mid y^{(1)}, \ldots, y^{(k)}) = D(\sigma(x) \mid \sigma(y^{(1)}), \ldots, \sigma(y^{(k)})),$$

*where $\sigma(x) := x_{\sigma(1)} \cdots x_{\sigma(d)}$.*

*We refer to the first condition as* $\oplus$-invariance *and to the second as* permutation invariance. *An operator sampling from a k-ary unbiased distribution is called a k-ary* unbiased variation operator.

1-ary (i.e., mutation only) operators are also called *unary* and we refer to 2-ary (i.e., crossover type) operators as *binary* ones. If arbitrary arity is considered, we call the corresponding model the $*$-ary unbiased black-box model.

A $k$-ary unbiased black-box algorithm [LW10] can now be described via the scheme of Algorithm 2. The *$k$-ary unbiased black-box complexity* of some class of functions $\mathcal{F}$ is the complexity of $\mathcal{F}$ with respect to all $k$-ary unbiased black-box algorithms.

Contrary to the unrestricted model, Lehre and Witt [LW10] could show that all functions with a single global optimum have a unary unbiased black-box complexity of $\Omega(n \log n)$, a bound which, for several standard test problems, is met by different unary randomized search heuristics, such as the $(1+1)$ EA or RLS. For results on higher arity models we refer to the work of Doerr et al. [DJK$^+$11].

**Ranking-Based Black-Box Models.** Another possible restriction of the black-box model was introduced by Doerr and Winzen [DW11]. The authors observed that many standard RSH do not

**Algorithm 2:** Scheme of a $k$-ary Unbiased Black-Box Algorithm

1. **Initialization:** Sample $x^{(0)} \in \{0,1\}^d$ uniformly at random and query $f(x^{(0)})$.
2. **Optimization:** for $t = 1, 2, 3, \ldots$ **until** *termination condition met* **do**
3.     Depending on $(f(x^{(0)}), \ldots, f(x^{(t-1)}))$ choose up to $k$ indices $i_1, \ldots, i_k \in [t-1]$ and a $k$-ary unbiased distribution $D(\cdot \,|\, x^{(i_1)}, \ldots, x^{(i_k)})$.
4.     Sample $x^{(t)}$ according to $D(\cdot \,|\, x^{(i_1)}, \ldots, x^{(i_k)})$ and query $f(x^{(t)})$.

take advantage of knowing *exact* objective values. Rather, for creating the next search points, many RSH always select those individuals with largest fitness values, examples are given below.

**Definition 2.** *Let $S$ be a finite set, let $f : S \to \mathbb{R}$ be a function, and let $C$ be a subset of $S$. The ranking $\rho$ of $C$ with respect to $f$ assigns to each element $c \in C$ the number of elements in $C$ with a smaller $f$-value plus 1, formally, $\rho(c) := 1 + |\{c' \in C \,|\, f(c') < f(c)\}|$.*

Note that two elements with the same $f$-value are assigned the same ranking.

Following [DW11], we restrict the two black-box models which we introduced in the previous section to black-box algorithms that use no other information than this ranking.

**Unrestricted Ranking-Based Black-Box Model.** The unrestricted ranking-based black-box model can be described via the scheme of Algorithm 1 where we replace the third line by "Depending on the ranking of $\{x^{(0)}, \ldots, x^{(t-1)}\}$ with respect to $f$, choose a probability distribution $p^{(t)}$ on $\mathcal{S}$."

**Unbiased Ranking-Based Black-Box Model.** For the definition of the unbiased ranking-based model we consider the scheme of Algorithm 2 and replace the third line by "Depending on the ranking of $\{x^{(0)} \ldots, x^{(t-1)}\}$ with respect to $f$, choose up to $k$ indices $i_1, \ldots, i_k \in [t-1]$ and a $k$-ary unbiased distribution $D(\cdot \,|\, x^{(i_1)}, \ldots, x^{(i_k)})$."

Both ranking-based black-box models capture many common search heuristics, such as $(\mu + \lambda)$ evolutionary algorithms, some ant colony optimization algorithms, and RLS. They do not include algorithms like simulated annealing algorithms, threshold accepting algorithms, or evolutionary algorithms using fitness proportional selection.

Doerr and Winzen [DW11] could show that while the basic and the ranking-based models yield the same asymptotic bounds for some problems (e.g., the ONEMAX function class), it does matter not to consider exact fitness values for other problems, e.g., the BINARYVALUE function class.

## 3 Minimum Spanning Trees

**Definition 3** (MST). *The* Minimum Spanning Tree *(MST) problem consists of a connected graph $G = (V, E)$ on $n := |V|$ vertices and $m := |E|$ weighted edges. The edge weights $w(e)$, $e \in E$, are positive real numbers. The objective is to find an edge set $E' \subseteq E$ of minimal weight that connects all vertices.*

We encode this problem in binary representation as follows. First, we enumerate the edges in $E$ in arbitrary order $\nu : E \to [m]$. For every bit string $x \in \{0,1\}^m$ we then interpret $x$ as the subset of edges $E_x := \{\nu^{-1}(i) \in E \,|\, x_i = 1\}$. In the following, we assume that the enumeration of the edges is *not known* to the algorithm. So the algorithm only knows the numbers $n$ and $m$, but neither knows the geometry of the graph nor which bit corresponds to which edge. However, it may assume that the graph is connected since otherwise no solution of MST exists.

For $E' \subseteq E$ let $c(E')$ be the number of connected components induced by $E'$, and let $w(E') = \sum_{e \in E'} w(e)$ be the total weight of $E'$. The book [NW10] argues that the objective function $f(E') = (c(E'), w(E'))$ is "appropriate in the black-box scenario". Thus, if an algorithm *queries* some $x \in \{0,1\}^m$, it receives $f(E_x) = (c(E_x), w(E_x))$ as answer.

In the ranking-based models, the objective value consists of a ranking of both components. That is, the oracle reveals two rankings of the search points, based on the first and the second component, respectively.

We obtain the following upper bounds by modifying Kruskal's algorithm to fit the black-box setting at hand.

**Theorem 4** (Upper Bounds for MST)**.** *The* (ranking-based) unrestricted *black-box complexity of the MST problem is at most* $2m+1$. *The* unary unbiased *black-box complexity is* $O(mn \log(m/n))$ *if there are no duplicate weights and* $O(mn \log n)$ *if there are. The* ranking-based unary unbiased *black-box complexity is* $O(mn \log n)$. *The* (ranking-based) binary unbiased *black box-complexity is* $O(m \log n)$. *The* (ranking-based) 3-ary unbiased *black-box complexity is* $O(m)$.

We conjecture that it is not an artifact of our methods that we obtain different upper bounds, but that all four complexity classes of the MST problem are different. The proofs can be found in the appendix, cf. Section B.

**Theorem 5** (Lower Bounds for MST)**.** *The unrestricted black-box complexity of MST for complete graphs is at least* $(1-o(1))n$.

*Proof.* We apply Yao's minimax principle [Yao77]. To this end, we show that there exists a probability distribution on the input set of all weighted complete graphs such that every deterministic algorithm needs at least $(1-o(1))n$ queries to compute a MST. More precisely, we consider the distribution $p$ on the set of all inputs where we sample uniformly at random a spanning tree, and give weight 1 to all of its edges. All other edges receive weight 2. We call edges of weight 1 "cheap", and all other edges "expensive".

Let us now consider a fixed deterministic algorithm $A$. We assume that the algorithm already knows which bit in the vector corresponds to which edge in the graph. This assumption makes life only easier for the algorithm. Then for each query the algorithm knows in advance how many connected component its query has. So the first component of the objective function does not contain any information. If the algorithm makes a query consisting of $k$ edges, then the total weight of all these edges is contained in the interval $[2k-n+1, 2k]$, depending of how many cheap edges the query contains. Therefore, each query gives at most $\log_2(n)$ bits of information.

Obviously, the algorithm $A$ needs to learn the set of all cheap edges. It is well known that the number of spanning trees on $n$ vertices is $n^{n-2}$ (so-calles Cayley's formula). Therefore $A$ needs to learn $(n-2)\log_2(n)$ many bits, so it has in the worst case a run time of $T := n-2$. Moreover, for every $0 \le t \le T$, after $T-t$ many queries the probability to find the correct solution is at most $n^{-t}$. Therefore, the probability that $A$ needs at least $T$ steps is at least

$$\Pr[T(I_p, A) \ge T] \ge 1 - \sum_{t=1}^{T-1} n^{-t} \ge 1 - \frac{n^{-1}}{1-n^{-1}} = 1 - o(1).$$

Note that the hidden constant in $o(1)$ does not depend on the algorithm $A$. By Markov's inequality, the expected run time is at least

$$\begin{aligned} \mathrm{E}[T(I_p, A)] &\geq T \cdot \Pr[T(I_p, A) \geq T] \\ &\geq T \cdot (1 - o(1)) \\ &= (1 - o(1))n. \end{aligned}$$

Since this holds for all deterministic algorithms $A$, Yao's minimax principle implies the statement. $\square$

In order to prove a lower bound in the unbiased setting, we compare MST with the auxiliary problem ONEMAX$_m$. The search space of ONEMAX$_m$ is the space $\{0, 1\}^m$, and for each vector $x \in \{0, 1\}^m$ the objective value is given by ONEMAX$_m(x) := \sum_{i=1}^{m} x_i$, the number of 1-bits in $x$.

**Theorem 6.** *The $k$-ary unbiased black-box complexity of MST for $n$ vertices and $m$ edges is at least as large as the $k$-ary unbiased black-box complexity of ONEMAX$_m$.*

*Proof.* For a given $m$, consider a path $P$ on $m+1$ vertices, all $m$ edges having unit weight. For the associated MST fitness function $f$ we have, for all bit strings $x \in \{0, 1\}^m$,

$$f(x) = (\text{OneMax}_m(x), m + 1 - \text{OneMax}_m(x)).$$

In particular, any algorithm solving optimizing $f$ can be used to optimize ONEMAX with the exact same number of queries. $\square$

It has been shown in [LW10] that ONEMAX$_m$ has a unary unbiased complexity of $\Theta(m \log m) = \Theta(m \log n)$ and in [DJK$^+$11] it was proven that the $*$-ary unbiased complexity of ONEMAX$_m$ is $\Theta(m/\log m) = \Theta(m/\log n)$. This yields the following.

**Corollary 7.** *The unary unbiased black-box complexity of MST is in $\Omega(m \log n)$; for all other arities, the unbiased black-box complexity of MST is in $\Omega(m/\log n)$.*

## 4 Single-Source Shortest Path

**Definition 8** (SSSP). *The* Single-Source Shortest Path *(SSSP) problem consists of a connected graph $G = (V, E)$ on $n := |V|$ vertices and $m := |E|$ edges. The edge weights $w(e)$, $e \in E$, are positive real numbers. There is a distinguished* source *vertex $s \in V$. The objective is to find for all vertices $v \in V$ a path $p_v$ in $G$ from $s$ to $v$ such that the total weight of $p_v$, $\sum_{e \in p_v} w(e)$, is minimal among all paths from $s$ to $v$.*

For the SSSP problem, it is less clear what a good choice of the search space and the objective function is. Two approaches have been regarded, which we discuss in the following subsections.

We assume without loss of generality that the nodes are labeled by $1, \ldots, n$ and that $s = 1$ is the source for which we need to compute the shortest path tree. Let $w : E \to \mathbb{R}^+$ be the weight function of the edges.

## 4.1 SSSP with Multi-Criteria Fitness

The paper [DJW06] argues for a multi-criteria objective function, where any algorithm may query arbitrary trees on $[n]$ and the objective value of any such tree is an $n-1$ tuple of the distances of the $n-1$ non-source vertices to the source $s = 1$ (if an edge is traversed which does not exist in the input graph, the entry of the tuple is $\infty$).

The paradigm underlying the unbiased black-box complexity is that the algorithm should not be allowed to exploit knowledge about solution candidates stemming from their representation, but only information stemming from their fitness and the population history. This is why only unbiased variation operators are admitted, which fulfill certain symmetry properties.

For the multi-objective formulation of the SSSP problem, we thus feel that there is little room for unbiasedness. With the fitness explicitly distinguishing the vertices, imposing certain symmetry conditions among the vertices makes little sense. A similar argument makes us not regard ranking-based black-box complexities for this problem.

[DJW06] shows that the unrestricted black-box complexity of this problem is lower bounded by $n/2$ and upper bounded by $2n - 3$.[3]

In this section, we first improve the bounds from [DJW06] and match them. Then we restrict the problem instances to complete graphs, which will avoid objective values of $\infty$ for the different objectives.

|       | arbitrary connected graph | complete graph |
|-------|---------------------------|----------------|
| upper | $n-1$                     | $\lfloor (n+1)/2 \rfloor + 1$ |
| lower | $n-1$                     | $n/4$          |

Table 3: Upper and lower bounds for the unrestricted black-box complexity of SSSP with multi-criteria objective function,

**Theorem 9.** *The unrestricted black-box complexity of SSSP with arbitrary input graphs is $n-1$.*

*Proof.* We start with the **upper bound**. We simulate Dijkstra's algorithm by first connecting all vertices to the source, then all but one vertices to the vertex of lowest distance to the source, then all but the two of lowest distance to the vertex of second lowest distance and so on, fixing one vertex with each query. This will cost an overall of $n-1$ queries.

For the **lower bound** consider the set $S$ of all graphs on $\{1, \ldots, n\}$ which contain exactly one path as edges (all of weight 1), one of the endpoints being the source $s = 0$, and no other edges. Again we apply Yao's minimax principle. To this end, let a deterministic algorithm $A$ be given; we show that $A$ uses in expectation $n-1$ queries on a graph drawn uniformly at random from $S$. We do this by showing that, in expectation, each query will give at most one new non-$\infty$ entry in the tuple. To this we shall apply the additive drift theorem for lower bounds [HY04].

We can assume without loss of generality that, during the run of algorithm $A$, the number of finite entries in the objective value does never decrease. Suppose that after some queries $A$ has determined $n - 1 - k$ finite entries in the objective value, so for $k$ vertices $A$ has still not discovered a path to the source. Let $T$ be the next query of $A$. Let $U$ be the set of all vertices that $A$ connects, in $T$, with a vertex with finite distance to the source. For all $v \in U$, let $t(v)$ be the number of vertices that $A$ connects to the source *via* $v$. The expected number of *new* non-$\infty$ entries in the objective value of $T$ is at most

$$\sum_{v \in U} \frac{1 + t(v)}{k}, \tag{1}$$

as $1/k$ is the probability that a given $v \in U$ is the next vertex in the path after the known vertices, and once we get that vertex right, we gain at most $1 + t(v)$ new non-$\infty$ entries. As we have a total of $k$ vertices left to connect with the source, we have $\sum_{v \in U} t(v) = k - |U|$. We have that (1) equals $|U|/k + (k - |U|)/k = 1$. Now the additive drift theorem [HY04] gives the desired bound. □

---
[3]Note that the upper bound holds in a restricted setting where the algorithm may only store up to two previous data points. However, the algorithm witnessing our upper bound for the unrestricted black-box setting does also not require the full storage granted by the unrestricted setting, but merely needs to store a linear number of pointers at any given time.

Surprisingly, if we may assume that the input graph is complete, we obtain a lower complexity. Note that this includes the case where the complete graph is obtained from on arbitrary one by adding dummy edges with artificially high weight. This shows, again, that even small changes in modeling the combinatorial problem can lead to substantial changes in the complexity.

For the upper bound, we cover the graph with $\lfloor (n+1)/2 \rfloor$ spanning trees, and query each of them. From the objective values, it is possible to compute all edge weights in the graph, and thus to compute the optimal solution. For the lower bound, we apply the same techniques as in the proof of 9, but use a more intricate distribution on the set of all spanning trees. Skipping the detais, we obtain the following.

**Theorem 10.** *The unrestricted black-box complexity of SSSP with* complete *input graphs is bounded from above by* $\lfloor (n+1)/2 \rfloor + 1$ *and bounded from below by* $n/4$.

*Proof.* We start with showing the upper bound. Essentially we prove that it is possible to learn the problem instance quickly.

We show that the complete graph $K_n$ on $n$ vertices may be written as the union of $\lfloor (n+1)/2 \rfloor$ spanning trees. If $n$ is even then it is well known that $K_n$ may be decomposed into $n/2$ edge-disjoint spanning trees [KK09]. If $n$ is odd, then we choose a node $v$ and all edges adjacent to $v$, thereby getting a spanning tree. The remaining edges form a $K_{n-1}$. Since $n-1$ is even we may decompose the remaining edges into $(n-1)/2$ spanning trees of $K_n - \{v\}$, which we complete to spanning trees of $K_n$ in an arbitrary way. Hence we have written $K_n$ as the union of $(n+1)/2$ spanning trees.

Now we describe our strategy. We choose a cover of $\lfloor (n+1)/2 \rfloor$ spanning trees as above. For each spanning tree $T$, we make a query to the oracle which contains exactly the vertices in $T$. After the query, we know for each vertex $v$ its distance from the source $s$. Since $T$ is a spanning tree, all distances are finite and all nodes can be reached via a unique path from $s$. Therefore, we can compute all the weights of edges in $T$. Since the spanning trees cover all edges, we know all edge weights after $\lfloor (n+1)/2 \rfloor$ queries. By our unrestricted computational power, we compute the minimal spanning tree and query for it.

As for the lower bound, we will use Yao's minimax principle (Theorem 18). We sample instances by having each vertex $i > 0$ choose uniformly at random a $j \in \{0, \ldots, i-1\}$ and then giving 1 as the weight to the edge between $i$ and $j$, and weight $n$ to all other edges. We call edges of weight 1 "cheap", and all other edges "expensive". By construction, the desired shortest path tree consists precisely of the the cheap edges (and for each $i$, the chosen $j$ is the predecessor on that shortest path tree).

Optimizing an instance as described above is only easier if we allow querying arbitrary sets of $n-1$ edges instead of trees, and even allowing these queries to be sequential (this means that the algorithm gets $n-1$ edge queries instead of one tree query). Each edge-query will reveal the weight of that edge to the algorithm. Furthermore, we will consider an algorithm done once all cheap edges have been queried in some previous query. Let a deterministic algorithm $A$ be given. For each $i \leq n$, $A$ has to find a needle-in-the-haystack of size $i$. Note that the different haystacks are completely independent; hence, the haystack associated with vertex $i$ requires an expected number of $(i+1)/2$ (edge) queries [DJW06]. It is of no importance in which order the algorithm uses its queries for the different haystacks. We can assume that all haystacks will be queried in order of increasing $i$, each until the cheap edge for vertex $i$ has been found. Thus, we get an overall expected number of

$$\sum_{i=1}^{n-1} \frac{i+1}{2} = \frac{n(n+1)}{4} - \frac{1}{2}.$$

*edge* queries; hence, $A$ will need an expected number $\geq n/4$ *tree* queries. □

We do not regard unbiased models for this setting. As mentioned before, it seems inappropriate to use an objective function that gives vertex-specific information and to still require unbiased variation operators.

## 4.2 SSSP with Single-Criterion Fitness

It is an interesting question how the bounds of Section 4.1 change if we require the algorithms to be unbiased. The unbiased model of Lehre and Witt in [LW10] has only been formulated for pseudo-Boolean functions, cf. Section 2. As we are dealing with a different representation here, our first step is to generalize the unbiasedness conditions to the setting of SSSP. As we discuss below, there is no unique best way to generalize unbiasedness to a more general class of problems. Even more, the upper bounds which we obtain for the different unbiasedness models differ by a factor of $\log n$, cf. Theorem 15 and Corollary 13. We conjecture that this difference is not an artifact of our analysis but that there actually exists an asymptotic difference between the two models.

In this section, we consider the following model for the SSSP problem. A representation of a candidate solution will be a vector $(\rho(2), \ldots, \rho(n)) \in [n]^{n-1}$ to be interpreted as follows. The predecessor of node $i$ is $\rho(i)$. Note that we do not require that $\rho(i) \neq i$, nor do we require that the candidate solution forms a tree; RSH without repair mechanisms might generate such solutions. In order to reflect the meaning of the components, the indices of such an $x$ will run from 2 to $n$, i.e., $x = (x_2, \ldots, x_n)$.

We can now formulate a first unbiasedness condition for this model. We require that, for any source-preserving permutation of the nodes, the probabilities are preserved. Intuitively, all subgraphs with the same structure but different labels are equally likely to be chosen.

**Definition 11.** *Let $k \in \mathbb{N}$. A $k$-ary structure preserving unbiased distribution is a family of probability distributions $\left(D(\cdot \mid y^{(1)}, \ldots, y^{(k)})\right)_{y^{(1)}, \ldots, y^{(k)} \in [n]^{n-1}}$ over $[n]^{n-1}$ such that for all inputs $y^{(1)}, \ldots, y^{(k)} \in [n]^{n-1}$ the distribution $D(\cdot \mid y^{(1)}, \ldots, y^{(k)})$ is invariant under relabeling of the non-source nodes.*

*That is, for all $\sigma \in S_n$ with $\sigma(1) = 1$ and for all $x \in [n]^{n-1}$ we have that*

$$D(x \mid y^{(1)}, \ldots, y^{(k)}) = D(\hat{\sigma}(x) \mid \hat{\sigma}(y^{(1)}), \ldots, \hat{\sigma}(y^{(k)})),$$

where $\hat{\sigma}(x) := \bigl(\sigma(x_{\sigma^{-1}(2)}), \ldots, \sigma(x_{\sigma^{-1}(n)})\bigr)$.

Alternatively, as search points are just mappings from the vertex set into itself, we might require that all possible images of each vertex are to be treated symmetrically. Formally, we require the following.

**Definition 12.** Let $k \in \mathbb{N}$. A family $\bigl(D(\cdot \,|\, y^{(1)}, \ldots, y^{(k)})\bigr)_{y^{(1)}, \ldots, y^{(k)} \in [n]^{n-1}}$ of probability distributions over $[n]^{n-1}$ is a *k-ary redirecting unbiased distribution* if, for all inputs $y^{(1)}, \ldots, y^{(k)} \in [n]^{n-1}$ the distribution $D(\cdot \,|\, y^{(1)}, \ldots, y^{(k)})$ is invariant under redirecting the nodes. That is, for all vectors $\sigma = (\sigma_2, \ldots, \sigma_n) \in S_n^{n-1}$ of permutations, and for all $x \in [n]^{n-1}$ we require

$$D(x \,|\, y^{(1)}, \ldots, y^{(k)}) = D(\vec{\sigma}(x) \,|\, \vec{\sigma}(y^{(1)}), \ldots, \vec{\sigma}(y^{(k)})),$$

where $\vec{\sigma}(x) := \bigl(\sigma_2(x_2), \ldots, \sigma_n(x_n)\bigr)$.

As in the hypercube model, we call an operator sampling from a $k$-ary structure preserving unbiased distribution a *k-ary structure preserving unbiased variation operator* and, similarly, an operator that samples from a $k$-ary redirecting unbiased distribution is called a *k-ary redirecting unbiased variation operator*.

To get a better understanding of the above definitions, let us investigate the restrictions for the unary case $k = 1$. We first look at the structure preserving model. Assume we have a search point $z$, and we would like to sample the next search point according to a probability distribution $D_z$ on the search space. We may do so if and only if there is an unbiased family of probability distributions $(D(\cdot \,|\, y))_y$ such that $D(\cdot \,|\, z) = D_z$. A necessary condition is

$$\text{For every source-preserving permutation } \sigma \text{ with } \hat{\sigma}(z) = z \text{ and all } x \in [n]^{n-1} \tag{2}$$
$$\text{it holds that } D_z(x) = D_z(\sigma(x)).$$

This condition is also sufficient. Assume $D_z$ satisfies (2). Then for every $y$ there are two possibilities.

1. Either there exists a $\sigma$ such that $\hat{\sigma}(y) = z$. In this case, we define $D(x \,|\, y) := D(\hat{\sigma}(x) \,|\, z)$ for all $x \in [n]^{n-1}$.

2. Or there does not exist such a $\sigma$. In this case, we let $D(\cdot \,|\, y)$ be the uniform distribution on the search space.

It can be checked that the family $D(\cdot \,|\, y)$ is a structure preserving unbiased family of distributions. The same holds for the redirecting unbiased model if we take $\sigma$ from the set of all vectors in $S_n^{n-1}$ and replace $\hat{\sigma}$ by $\vec{\sigma}$.

Now we can determine the distributions $D_z$ satisfying condition (2). For the structure-preserving model, we need to find all source-preserving permutations that leave $z$ unchanged. These are in one-to-one correspondence with the source-preserving automorphisms of the graph induced by $z$. If $A$ denotes the group of these automorphisms, then condition (2) is that $D_z$ must be invariant under $A$, i.e., for all $\alpha \in A$ and all $x \in [n]^{n-1}$ we require $D_z(x) = D_z(\alpha(x))$.

For the redirecting model, we need to determine all families $\sigma \in S_n^{n-1}$ such that $\vec{\sigma}(z) = z$. Consider any component $y_i$ of $y$. Then we can choose $\sigma_i$ to be any permutation of $[n]$ with $\sigma_i(z_i) = z_i$. In particular, for all $s, t \in [n] \setminus \{z_i\}$ there is such a permutation mapping $s$ to $t$. Therefore, a distribution $D_z$ is redirecting unbiased if and only if the following condition is satisfied. 'If $x^{(1)}, x^{(2)} \in [n]^{n-1}$ are vectors such that for every $i \in [2, \ldots, n]$ the equations $x_i^{(1)} = z_i$ and $x_i^{(2)} = z_i$ are either both true or are both false, then $D_z(x^{(1)}) = D_z(x^{(2)})$." Similar considerations hold for higher arity. We omit the details.

Since we do not want the objective function to give vertex-specific information, we use the single-criterion objective function $f_G(\rho(2), \ldots, \rho(n)) := \sum_{i=2}^{n} d_i$ where $d_i$ is the distance of the $i$-th node to the source. If an edge – including loops – is traversed which does not exist in the input graph, we set $d_i := C$ where $C$ is some very large constant (e.g., we could choose $C := nw_{\max}$). In all models, the constant $C$ can be learned by the algorithm in a constant number of queries, e.g., by querying the objective value of search point $(2, \ldots, 2)$ in the unrestricted model and dividing it by $n-1$ and, similarly, querying a search point with "all nodes to one non-source node" in the structure preserving unbiased model, and "all nodes to the same node" in the redirecting unbiased model. In the latter one we may perform two independent such queries to guarantee a probability of at least $1 - n^{-2}$ to learn the exact value of $C$. Thus, we assume that the value $C$ is known to the algorithm.

It has been argued in [BBD+09] that the RLS algorithm, which in each iteration flips exactly one bit chosen uniformly at random, solves the single-source shortest path problem with the single-criterion objective function in $O(n^3)$ iterations. Since this algorithm is contained in the ranking-based unrestricted black-box model, we immediately gain an upper bound of $O(n^3)$. RLS is also contained in the redirecting unbiased model.

**Corollary 13.** *The ranking-based unrestricted black-box complexity and the ranking-based unary redirecting unbiased black-box complexity of the SSSP with the single-criterion fitness function is $O(n^3)$.*

We conjecture that already for the (non-ranking-based) unary redirecting unbiased model this bound is tight. However, the following shows that we can achieve better bounds in the unrestricted black-box model.

**Theorem 14.** *The unrestricted black-box complexity of the SSSP with the single-criterion objective function is at most $\sum_{i=1}^{n-1} i = n(n-1)/2$ and the ranking-based unrestricted one is at most $(n-1)^2$.*

*Proof.* Let $G = (V, E)$ be a connected graph and $w$ a positive weight function on $E$. We show that adding the $i$-th node to the shortest-path tree costs at most $n-i$ queries in the unrestricted model and at most $n-1$ queries in the ranking-based unrestricted model. Basically, we are imitating Dijkstra's algorithm. We say that a node is *unconnected* if in the current solution there does not yet exist a path from that node to the source and we say it is *connected* otherwise. Recall that the indices run from 2 to $n$, so $(2, 3, \ldots, n)$ encodes the graph where every node except the source points to itself and is thus not connected to the source.

In the first $n-1$ iterations query the strings $(1, 3, 4, \ldots, n)$, $(2, 1, 4, 5, \ldots, n)$, ..., $(2, \ldots, n-1, 1)$, each of which connects exactly one node to the source and lets all other node point to themselves. Then each of these strings has $n-2$ unconnected nodes, which contribute equal to the fitness function. In the non-ranking-based model, we learn the costs of the edges adjacent to the source. In the ranking-based model, we still learn their ranking. In particular, in both models we learn which node can be connected to the source at the lowest cost.

Now assume that $k < n-1$ nodes $v_1, \ldots, v_k$ have been added to the shortest path tree already. We call the remaining non-source nodes "free nodes".

In the non-ranking-based model, we proceed as follows. Test all $n-k-1$ possibilities to connect exactly one free node to $v_k$ and let every other free node point to itself. We learn the costs of all edges between free nodes and $v_k$. Furthermore, we compute for each $j < k$ the lowest cost for connecting a free node to the source via $v_j$. Note that for $j < k$ we have gathered the required information in previous steps. The cheapest such connection is added to the current solution, and we denote this node by $v_{k+1}$. Thus, we have constructed the shortest path tree in $\sum_{i=1}^{n-1} i = n(n-1)/2$ queries.

In the ranking-based model we perform the following $n-1$ queries in the $k$-th step. In each query, we connect the node $v_1, \ldots, v_k$ as learned before. For the free nodes, we query the following combinations.

- For each free node $v$ make the following query. Connect $v$ to $v_k$, and let all other free nodes point to themselves.

- For each $j \in [k-1]$, take the free node with minimal edge cost to $v_j$. Connect this node to $v_j$, and connect all other free node to themselves.

Note for the second type that we know the free node with minimal edge cost to $v_j$ because we have learned the ranking of all edges adjacent to $v_j$ in an earlier step.

Since all queries have exactly $n - k - 1$ unconnected nodes, the contribution of these nodes to the cost function is equal for all queries. Thus we learn which free node produces minimal cost when attached to $n_1, \ldots, n_k$. We call this node $v_{k+1}$. Moreover, we learn the ranking of all edges from free nodes to $v_k$, which we need in the forthcoming steps.

Together, the algorithm adds an additional vertex to the current solution using $n-1$ queries. Since we have to add $n-1$ vertices in total, the claim follows. □

This theorem, as all the subsequent theorems, can be proven by applying some variants of Dijkstra's algorithm. It is possible to derive the edge weights in the unrestricted model from the oracle's answers, so we need only $n-1-i$ queries to add the $i$-th vertex to the tree. In the ranking-based model we have less information and need up to $(n-1)$ queries to add a new edge.

For the structure-preserving unbiased model, things get more involved. In the unary case we need $O(n^2 \log n)$ queries to add a new edge. In the binary and 3-nary case, some precomputations are possible that reduce the run time.

**Theorem 15.** *The unary structure-preserving unbiased black-box complexity of SSSP is $O(n^3 \log n)$.*

*Proof.* In the structure-preserving unbiased black-box we are not allowed to direct a free node[4] to the node that was lastly added to the search tree. For this reason, in each step we need not only to search for the node that we want to add, but also for the place where we want to attach it.

To add the first node, the algorithm does the following. It queries $3n \log n$ times a vertex using the following operator. Draw $x \in [n]^{n-1}$ such that $x$ has exactly one entry $j$ equaling 1 (i.e., that is connected to the source) and such that all every other entry points to itself. Clearly, all but one node are connected to the source and all other have contribution $C$ to the costs.

By standard coupon collector arguments we have that the probability to have connected each possible edge at least once is at least $1 - n^{-2}$. To continue, let $x$ be one of the queried search points with the lowest costs. Then $x$ encodes a shortest path.

It is lengthy, but straightforward to verify that the above variation operator is a (0-ary) structure-preserving unbiased one. We omit this computation.

---
[4] Again, we say that a node is free if its shortest path to the source is not yet known.

Assume now that we have already learned $k$ nodes $v_1, \ldots, v_k$ to be added to the shortest path tree. We do not assume that we have a search point $x$ encoding the whole shortest path tree. However, we require that for every $j \in [k]$, we have a search point encoding the shortest path to $v_j$. Let $x^k$ be the search point encoding the shortest path to $v_k$, the vertex that we have discovered latest. We query $3k(n-k)\log(k(n-k))$ times the following. "Create the path $z$ from $x^k$ by pointing exactly one of the unconnected nodes to one of the $v_j$, for $j \in [k]$. Let every other unconnected node point to itself."

With probability at least $1 - n^{-2}$ we have queried all $n - 1 - k$ possible attachments of free nodes to an $n_j$. We compute the lowest cost and continue with one of the vertices sampled in the last phase with lowest cost.

It is again lengthy, but straightforward to verify that the variation operators is structure-preserving unbiased. We omit the details.

The complexity of the problem is upper bounded by the run time of the algorithm, which is

$$O\left(\sum_{k=1}^{n} k(n-k)\log(k(n-k))\right) = O(n^3 \log n).$$

$\square$

**Theorem 16.** *The binary structure-preserving unbiased black-box complexity of SSSP is $O(n^2 \log n)$.*

We give an intuitive proof of this theorem here; a detailed analysis can be found in the appendix, in Section C.

*Proof(sketch).* We imitate Dijkstra's algorithm. Note however, that in the structure-preserving unbiased model we are not allowed to (i) direct a node to some node of our choice, e.g., to the node that was lastly added to the search tree; and (ii) we cannot simply add a vertex to the current solution but need to construct this new solution.

To overcome the first point, we split up the algorithm in two phases. In the **search phase**, we do not actually find a search point encoding the search tree $T$, but rather for every leaf $v$ of the tree we store a search point that contains the path in $T$ from the source to $v$, with all other nodes pointing to themselves.

When the search phase is completed, we know the structure of the search tree, and start the **construction phase**. In this phase, we grow the desired search tree $T$.

Remarkably, we need binary operators only to check whether we have added the correct vertex to the tree in the construction phase.

Let us start with the search phase. For the first step, let $v_1$ be the source, and store the search point where every vertex points to itself. For learning how to add the $k+1$-th node, assume that for $k$ nodes $v_1, \ldots, v_k$ we have learned already how to add them to the shortest path tree. We call the other nodes "free nodes".

Consider the search point $x^{(k)}$ lastly found. As mentioned above, $x^{(k)}$ consists of a the shortest path from the source to $v_k$, and all nodes not on this path point to themselves. We call the nodes on the shortest path from $v_1$ to $v_k$ "connected nodes". Note that a node cannot at the same time be free and connected, but there may be nodes that are neither free nor connected. We apply $O(n \log n)$ times the following unary unbiased operator. "Create the path $z$ from $x^{(k)}$ by pointing exactly one of the unconnected nodes to $v_k$. Let every other unconnected node point to itself."

The operator is unbiased since every source-preserving automorphism of $x^{(k)}$ must fix the path in $x^{(k)}$, and hence must fix $v_k$.

With high probability we have queried all possible attachments of free nodes to $v_k$. We compute the lowest cost for connecting one of them via $v_k$ and compare it with the lowest cost for connecting a free node via one of the vertices $v_1, \ldots, v_{k-1}$. Of all these connections, we choose the cheapest and store the corresponding search point.

Since we need to add $n - 1$ vertices, the search phase needs $O(n^2 \log n)$ queries.

For the construction phase, assume that we have learned the complete shortest path tree $T$. I.e., for every vertex $v$, we have a search point encoding the path in $T$ from the source to $v$. We want to construct $T$ explicitly. We start with the empty search tree (i.e., the search point where all nodes point to themselves), and add vertices in a depth-first manner.

Now we describe how to construct $T$ iteratively. We call the queries between adding the $(k-1)$st and $k$-th node the *k-th phase*. At the end of each phase, we choose an *active node*, to which the next node is to be attached. We start with the source being active.

For the $(k+1)$st phase, assume we have already constructed a tree $T_k$ of size $k$, encoded in a search point $y^{(k)}$. We will make use of the unary operator $\texttt{attach}(y^{(k)})$, which creates a search point $z$ from $y^{(k)}$ by redirecting exactly one unconnected node to the active node. In general, this operator does not need to be unbiased, since there may be automorphisms of the search tree mapping the active node somewhere else. However, it is possible to avoid such automorphisms by carefully choosing the order in which the depth first search traverses the children of the active node. The key idea is to traverse longest search paths first. We omit the details.

Let $v$ be the node we want to attach to the active node, and let $x_v$ be the search point storing the shortest path from $s$ to $v$, i.e., $x_v = x^{(i)}$ for some $i \in [n]$. We add $v$ to $T_k$ as follows. Sample $z \leftarrow \texttt{attach}(y^{(k)})$. Skipping the details, we note that it is possible by binary unbiased operations to compute in a constant number of queries the number of edges in which $z$ and $x_v$ coincide. Hence we can decide whether the newly added node was $v$. We keep on sampling $z \leftarrow \texttt{attach}(y^{(k)})$ until we find a $z$ that adds $v$ to $y^{(k)}$, then setting $y^{(k+1)} \leftarrow z$. If $v$ has a child in $T$ then we make $v$ the next active node. Otherwise, we backtrack until we find a node that has an unconnected child, and make this node active.

The expected number of queries needed to add a new node is $O(n)$. Since $n$ nodes need to be added, the construction phase needs $O(n^2)$ queries. □

By allowing 3-ary operators it is possible to imitate Dijkstra's algorithm more directly, without any need to split up the algorithm into two phases as in the proof of the previous theorem. We get the following theorem.

**Corollary 17.** *The 3-ary structure-preserving unbiased black-box complexity of SSSP is $O(n^2)$.*

## 5 Conclusions

This first analysis of the different black-box complexity notions for two classic combinatorial optimization problems showed the following. In general, all notions make sense for combinatorial problems as well, though some care has to be taken of how to implement unbiasedness conditions. The particular bounds we find are reasonably close to actual run times observed by existing randomized search heuristics, that is, the black-box complexities give reasonable bounds for the problem difficulties here. In cases where our bounds are smaller than those observed by current best search heuristics, further studies are needed to determine whether current heuristics can be improved, e.g., by using higher-arity variation operators, or whether additional restrictions to the black-box model are needed to exclude artificial algorithms.

## Acknowledgment

Timo Kötzing was supported by the Deutsche Forschungsgemeinschaft (DFG) under grant NE 1182/5-1.

Carola Winzen is a recipient of the Google Europe Fellowship in Randomized Algorithms, and this research is supported in part by this Google Fellowship.

# Appendix

## A  Mathematical Background

The following presentation of Yao's minimax principle is taken from the book by Motwani and Raghavan [MR97].

**Theorem 18** (Yao's Minimax Principle). *Let $\Pi$ be a problem with a finite set $\mathcal{I}$ of input instances (of a fixed size) permitting a finite set $\mathcal{A}$ of deterministic algorithms. Let $p$ be a probability distribution over $\mathcal{I}$ and $q$ be a probability distribution over $\mathcal{A}$. Then,*

$$\min_{A \in \mathcal{A}} \mathrm{E}[T(I_p, A)] \leq \max_{I \in \mathcal{I}} \mathrm{E}[T(I, A_q)],$$

*where $I_p$ denotes a random input chosen from $\mathcal{I}$ according to $p$, $A_q$ a random algorithm chosen from $\mathcal{A}$ according to $q$ and $T(I, A)$ denotes the running time of algorithm $A$ on input $I$.*

Furthermore, we will use the following *drift theorem* in our proofs.

**Theorem 19** (Additive Drift [HY04]). *Let $(X_t)_{t \geq 0}$ be random variables describing a Markov process over a finite state space $S \subseteq \mathbb{R}$. Let $T$ be the random variable that denotes the earliest point in time $t \geq 0$ such that $X_t = 0$. If there exist $c > 0$ such that*

$$E[X_t - X_{t+1} | T > t] \leq c,$$

*then*

$$E[T | X_0] \geq \frac{X_0}{c}.$$

*If there exist $d > 0$ such that*

$$E[X_t - X_{t+1} | T > t] \geq d,$$

*then*

$$E[T | X_0] \leq \frac{X_0}{d}.$$

Rather than bounding the expected runtime of an algorithm, it is sometimes easier to show that it solves the given problem with good probability in some number $s$ of iterations. If we are only interested in asymptotic black-box complexities, the following remark allows us to use such statements for upper bounds.

**Remark 20.** *Suppose for a problem $P$ there exists a black-box algorithm $\mathcal{A}$ that, with constant success probability, solves $P$ in $s$ iterations. Then the black-box complexity of $P$ is at most $O(s)$.*

*Proof.* Let $c$ be an upper bound for the failure probability of algorithm $\mathcal{A}$ after $s$ iterations. We call the $s$ iterations of $\mathcal{A}$ a *run* of $\mathcal{A}$. If $X_i$ denotes the indicator variable for the event that the $i$-th independent run of $\mathcal{A}$ is successful (i.e., computes an optimum), then $\Pr[X_i = 1] \geq 1 - c$. Clearly, $Y := \min\{k \in \mathbb{N} \mid X_k = 1\}$ is a geometric random variable with success probability at least $1 - c$. Hence, $\mathrm{E}[Y] = (1 - c)^{-1}$, i.e., the expected number of independent runs of $\mathcal{A}$ until success is at most $(1 - c)^{-1}$. Thus, we can optimize $P$ in an expected number of at most $(1 - c)^{-1} s$ iterations. Since $c$ is at least constant, the claim follows. □

# B  Upper Bounds for MST

In this section, we prove Theorem 4.

We start with the statement in the unrestricted setting, which is very simple.

*Proof for Unrestricted Black-Box Complexity.* Query the empty graph $(0,\ldots,0)$ as a reference point, then query all edges $e_i$, $i \in [m]$, where $e_i$ denotes the $i$-th unit vector $(0,\ldots,0,1,0,\ldots,0)$. Then test all edges in increasing order of their weights (ties broken arbitrarily). Accept an edge if it does not form a cycle (note that this can be checked trough the first component of the bi-objective objective function). □

The unbiased model is much more involved. For all three arities, the basic principle of the algorithm is the same as for the unrestricted algorithm, basically following Kruskal's algorithm for MST construction. In the **first step**, we create the empty graph. This serves as a reference point for all further iterations.

In the **second step**, we learn (in the unbiased model) or order (in the ranking-based model) the weights of the edges (including multiplicities) and we test the inclusion of the edges in increasing order of their weights in the **third step**. From basic facts about Kruskal's algorithm we know that for edges with the same weight, it does not matter in which order we test them.

In the following, we prove upper bounds for the expected number of queries needed to complete each step. For readability purposes, we split the proof into 4 parts, one for each model.

Let us remark already here that in the proof we apply only few variation operators, namely `uniform()` which samples a bit string $x \in \{0,1\}^n$ uniformly at random, $\text{RLS}(\cdot)$ (random local search) which, given some $x \in \{0,1\}^n$, creates from $x$ a new bit string $y \in \{0,1\}^n$ by flipping exactly one bit in $x$, the bit position being chosen uniformly at random. We also use the operator $\text{complement}(\cdot)$, which assigns to every $x \in \{0,1\}^n$ its bit-wise complement $\bar{x}$. Lastly, we use the operator $\text{RLS}_k(\cdot,\cdot)$. Given some bit stings $x,y \in \{0,1\}^n$, $\text{RLS}_k(x,y)$ outputs a bit string $z$ that has been created from $x$ by flipping exactly $k$ bits of $x$, chosen uniformly at random, in which $x$ and $y$ differ. If $x$ and $y$ differ in less than $k$ bits, it outputs $x$.

The following is straightforward to verify, for a discussion of the properties of unbiased variation operators confer [LW10].

**Remark 21.** `uniform()` *is a (0-ary) unbiased variation operator and both* $\text{RLS}(\cdot)$ *and* $\text{complement}(\cdot)$ *are unary unbiased ones. Furthermore, for all $k$, $\text{RLS}_k(\cdot,\cdot)$ is a binary unbiased variation operator.*

## B.1  The Unary Unbiased Model

The bound of $O(mn \log m)$ for multiple edge weights will follow from the bound for the ranking-based unary unbiased model, so we postpone it to section B.2.

*Proof of $O(mn \log(m/n))$ bound.*

**First step.** As required by the unbiased black-box model, we first draw a search point $x \in \{0,1\}^m$ uniformly at random. We construct the empty graph by creating $y \leftarrow \text{RLS}(x)$ and accepting $x \leftarrow y$ if and only if $w(y) < w(x)$. We do so until $w(x) = 0$. By the standard coupon collector argument, this takes an expected $O(m \log m)$ queries. In the following, we denote the empty graph by $x^0$.

**Second step.** In order to learn the weights, we again employ the operator $\text{RLS}(\cdot)$ iteratively to $x^0$ until we have added all edges. More precisely, we generate a sequence of search points $x^k$ as follows. In the $k$-th iteration of the second step we create $z \leftarrow \text{RLS}(x^{k-1})$ and query the objective value of $z$. If it is larger than the objective value of $x^{k-1}$, we set $x^k \leftarrow z$. Otherwise, we discard $z$ and keep on sampling from $x^{k-1}$. Then the difference of the objective values of $x^k$ and $x^{k-1}$ is exactly the weight of the edge added in the $k$-th iteration, so we learn all edge weights. By the same coupon collector argument as before, this takes an expected $O(m \log m)$ queries. Let $w_1 \leq \ldots \leq w_m$ be the ordering of the weights.

**Third step.** We return to the search point $x^0$, and show how to construct an MST for the underlying graph. For any $k \leq n-1$ we call the queries needed to add the $k$-th edge to the MST the $k$-th phase. We start by applying $\text{RLS}(\cdot)$ to $x^0$ until we have found an edge of minimal weight $w_1$. We call the last sample $y^1$.

Assume now that we have already added $i$ edges of the MST. Let $y^i$ be this search point and let us assume that we have tested $t_i$ edges to find the $i$-th one. To test the inclusion of the $(t_i + 1)$st edge, we query $z^{(i,t)} \leftarrow \text{RLS}(y^i)$. By the change in the second value of the objective function $w(z^{(i,t)}) - w(y^i)$ we learn how large the weight of the flipped edge is and by the first value $c(z^{(i,t)})$ we learn whether we can add this edge without creating a cycle (if and only if $c(z^{(i,t)}) < c(y^i)$).

We do so until we have flipped the $(t_i+1)$st heaviest edge (or one of them, if there exist multiple such edges). This requires an expected number of $m$ queries. If we cannot add this edge to the current solution without creating a cycle, we check whether we have already created in one of the $z^{(i,t)}$s the string which includes the $(t_i + 2)$nd heaviest edge. If so, we check whether or not to add it to the current solution. If we have not created it already, we continue drawing $z^{(i,t)} \leftarrow \text{RLS}(y^i)$ until we have found the edge with the lowest weight that can be included to our current solution $y^i$. We call the new solution $y^{i+1}$ and continue with the $(i+2)$nd phase until we have added a total number of $n-1$ edges to $x^0$.

To determine an upper bound for the number of queries needed, let $k_i := t_i - t_{i-1}$ be the number of edges for which we have tested the inclusion in the $i$-th phase (including the lastly included one). By a coupon collector argument the expected number of queries needed in the $i$-th phase is $m/k_i \cdot k_i \log k_i = m \log k_i$. This follows from the fact that we need to sample all $k_i$ edges ("coupons") but the chance of getting one of them equals only $m/k_i$, for each query. Note that this argument works only in the case when all edge weights are distinct. Otherwise, we would need to sample more often to be sure that we have seen all edges of the given weight.

This shows that the third phase of the algorithm takes no more than $O(m \sum_{i=1}^{n-1} \log k_i)$ queries. Since $\sum_{i=1}^{n-1} k_i \leq m$ we conclude that

$$\sum_{i=1}^{n-1} \log k_i = \log \big( \prod_{i=1}^{n-1} k_i \big) \leq \log \big( (m/n)^n \big) = n \log(m/n),$$

and thus, $O(m \sum_{i=1}^{n-1} \log k_i) = O(mn \log(m/n))$ Hence, the unary unbiased black-box complexity of MST is

$$O(m \log m) + O(m \log m) + O(mn \log(m/n)) = O(mn \log(m/n)).$$

□

## B.2 The Ranking-Based Unary Unbiased Model

In the following, we speak of *weights*, even if we are in the ranking-based black-box model. Note however, that we do not need to know the exact value of the weight but only its *rank*.

*Proof of $O(mn \log n)$ bounds.*

**First step.** The first step is exactly the same as in the unary unbiased model. Note that, thanks to the information given by the ranking, we are still able to determine if $w(y) < w(x)$ holds. Again we denote the empty graph by $x^0$.

**Second step.** For this model, we can skip the second step.

**Third step.** The difference for the ranking-based unbiased model compared to the unbiased one is quite obvious. Since we cannot query for the objective value $(c(x), w(x))$ but only the relative ranks of $c(x)$ and $w(x)$, we do not know which bits we have flipped in either one of the iterations. Thus, for the inclusion of the $(i+1)$st edge, we perform $O(m \log m)$ queries $z^{(i,t)} \leftarrow \texttt{RLS}(y^i)$ to find, with high probability, the edge with the smallest weight that can be included into the current solution. Note that we can check the ranking of the weights via the second component of the objective function and the feasibility of adding it to the current solution via the first component. If and only if the rank of $c(z^{(i,t)})$ is strictly less than that of $c(y^i)$ we can include the corresponding edge. Since we need to include $n-1$ edges into the MST, we need an expected $O(nm \log m)$ queries until all edges of the MST have been added. □

## B.3 The Ranking-Based Binary Unbiased Model

Compared to the unary case, the binary unbiased black-box complexity allows to gain information about the Hamming distance for two search points. We formalize this in the following lemma.

**Lemma 22.** *There is a procedure using binary unbiased variation operators only such that, given $k$, $x$ and $y$, decides whether the Hamming distance from $x$ to $y$ is $k$ (on MST objective functions).*

*Proof.* We define a binary unbiased operator which returns $x$ if the distance between $x$ and $y$ is $k$ and otherwise returns a 1-Hamming neighbor of $x$. Due to the nature of MST objective functions, a 1-Hamming neighbor of $x$ cannot have the same objective value as $x$ (an edge was either added or removed, and all weights of edges are positive). □

*Proof of $O(m \log n)$ bound.*

**First step.** As required by the unbiased black-box model, we first draw a search point $x \in \{0,1\}^m$ uniformly at random.

We can create the empty graph in an expected time of $2m$ as follows. Set $y \leftarrow \texttt{complement}(x)$. We then set $z \leftarrow \texttt{RLS}_1(x, y)$ with probability $1/2$ and $z \leftarrow \texttt{RLS}_1(y, x)$ with probability $1/2$. We update $x \leftarrow z$ (in the first case) and $y \leftarrow z$ (in the second case) if and only if $w(z) < w(x)$ or $w(z) < w(y)$, respectively.

With probability $1/2$, this operation decreases the Hamming distance of $x$ and $y$ by 1 and thus, the expected number of calls to $\texttt{RLS}_1(\cdot, \cdot)$ is $2m$ (just note that initially $x$ and $y$ had a Hamming distance of $m$). Note that choosing $\texttt{RLS}_1(x,y)$ with probability $1/2$ and $\texttt{RLS}_1(y,x)$ with probability $1/2$ is still an unbiased operation. As before, let $x^0$ denote the empty graph.

**Second step.** The binary model allows us to store the information which bits have been flipped already, where we need time $O(\log m)$ to add a bit to the storage, and where we can perform a lookup, i.e., decide whether a bit has already been stored, in constant time. The idea is to maintain a search point $s$ consisting of all the edges that have been stored. Adding an edge is done by binary search, and given a graph consisting of only a single edge, the Hamming distance of $s$ tells us whether the edge is stored or not (here we use Lemma 22).

With this tool, we can accelerate the second phase. We query an expected number of $m \log m$ times the objective value of $x^t \leftarrow \text{RLS}(x^0)$ until we have learned all $m$ different weights/rankings. For each sample, we lookup whether it has been stored. If so, it is discarded. If not, its edge is added to the storage. We need to perform $m$ store operations and $O(m \log m)$ lookups, both contribution $O(m \log m)$ to the runtime. Therefore, we can learn all weights with multiplicities in time $O(m \log m)$.

We fix some labeling of the the edges $e_{j_1}, \ldots, e_{j_m}$ such that their weights are ordered $w_1 = w(e_{j_1}) \leq \ldots \leq w_m = w(e_{j_m})$. Note that for every $i$, we have already sampled a search point $y^i$ containing only the edge $e_i$.

**Third step.** As in the unary model, we successively add edges to our current solution. This time, we call the queries needed to *test* the inclusion of the $i$-th heaviest edge $e_{j_i}$, given that we have tested already the inclusion of edge $e_{j_{i-1}}$, the *$i$-th phase*.

We show that such a phase requires at most $O(\log n)$ queries. Let $x$ be the current search point. Let $d$ be the Hamming distance from $x$ to the search point $y^i$ containing only the edge $e_i$. For simplicity, assume that $d$ is even. Then we flip exactly half of the bits in which $x$ and $y^i$ differ by setting $z \leftarrow \text{RLS}_{d/2}(x, y^i)$. By computing the Hamming distance between $z$ and $x^0$, we may decide whether we have flipped $e_i$ or not. If $e_i$ is not contained in $z$ then we discard $z$ and try again. Otherwise, we replace $y^i$ with $z$ and again flip half of the bits in which $z$ and $x$ differ, updating $z$ whenever $e_i$ is contained in the new sample.

Repeating like this, we have in each step a probability of $1/2$ to reduce the Hamming distance between $z$ and $x$ by half. Meanwhile, by comparing with $x^0$ we ensure that $e_i$ is always contained in $z$. Therefore, when the Hamming distance of $x$ and $z$ decreases to 1, the search point $z$ differs from $x$ only by the edge $e_i$, as desired.

As for the ranking-based unary unbiased model, the objective value of $z$ tells us whether to include the edge $e_i$ or not. For the $i$-th phase, we need $O(\log n)$ queries. As we need to check each edge, the third step needs time $O(m \log n)$. □

## B.4 The Ranking-Based 3-Ary Unbiased Model

In the 3-ary model, we have some more flexibility. The main advantage is that we can create any particular 1-bit flip in a linear number of queries (linear in the length of the bit string). Using this, we can optimize the MST problem in a linear number of queries. Again, we use the word "weight" but are aware that we are not given the exact value but only its rank.

*Proof of $O(m)$ black-box complexity.* **First step.** The bound for the binary model holds for the 3-ary as well, and thus, $O(m)$ is an upper bound for the first step in the 3-ary model, too. Let us again denote the empty graph by $x^0$.

**Second step.** We show how to learn the weights of the edges in linear time. To encode which edges have been looked at already, we set $y \leftarrow \text{complement}(x)$, the bit-wise complement of $x$. Throughout the run of the algorithm it will hold that the edges we have looked at correspond to the bit positions in which $x$ and $y$ do not differ.

We learn the first edge weight by querying $z \leftarrow \text{RLS}(x^0)$. We update $y \leftarrow \text{update}(y, z, x^0)$, where $\text{update}(\cdot, \cdot, \cdot)$ is the 3-ary variation operator that can be described as follows. Given $a, b, c \in \{0, 1\}^m$, the operator $\text{update}(a, b, c)$ returns $c$ for those positions where $a$ and $b$ do not differ and returns $a$ otherwise.

It is easy to verify that the Hamming distance of $y$ and $x^0$ decreases by 1 in each such step. Furthermore, we have created all possible 1-bit flips, after only $m$ such steps (i.e., $2m$ queries since each step consists of 2 queries). Let us again fix an ordering of the edges $e_1, \ldots, e_m$ such that $w(e_1) \leq \ldots \leq w(e_m)$. Let $z_1, \ldots, z_m \in \{0,1\}^m$ be the corresponding ordering of the bit strings which are of Hamming distance 1 to $x^0$ (the $z$s in the description above).

**Third step.** We now check the inclusion of the edges to the current solution in increasing order of their weights. Since $e_1$ can be included without any further consideration, we may assume that we have already tested the inclusion of the $i$ edges with the lowest weights. Let $x$ be our current solution. To test the inclusion of edge $e_{i+1}$ into the current solution, we query $z \leftarrow \texttt{test}(x, x^0, z_{i+1})$, which is again an unbiased 3-ary variation operator that can be described as follows. For any $a, b, c \in \{0,1\}^m$ the operator $\texttt{test}(a, b, c)$ outputs a string that has entries equal to $a$ in all positions for which $b$ and $c$ do not differ and entries equal to $1 - a$ otherwise.

Note that in our case, we clearly have that the strings $x$ and $z$ differ in exactly one bit position (because $x^0$ and $z_{i+1}$ do). We update $x \leftarrow z$ if the edge $e_{i+1}$ can be included into the current solution and we continue with testing the inclusion of edge $e_{i+2}$ otherwise.

As this third phase requires at most $m$ queries, the ranking-based unbiased 3-ary black-box complexity of MST can be bounded by $O(m)$. □

## C Single-Criterion SSSP

In this section, we carry out the details for the proof of Theorem 16.

*Proof.* We imitate Dijkstra's algorithm. Note however, that in the structure-preserving unbiased model we are not allowed to (i) direct a node to some node of our choice, e.g., to the node that was lastly added to the search tree; and (ii) we cannot simply add a vertex to the current solution but need to construct this new solution.

To overcome the first point, we split up the algorithm in two phases. In the **search phase**, we do not actually find a search point encoding the search tree $T$, but rather for every leaf $v$ of the tree we store a search point that contains the path in $T$ from the source $s$ to $v$, with all other nodes pointing to themselves.

When the search phase is completed, we know the structure of the search tree, and start the **construction phase**. In this phase, we actually grow the desired search tree.

Let us start with the search phase. In the first step, we set $v_1 := s$ and store the search point where every vertex points to itself. For the $k+1$-th step, assume that we have already learned $k$ nodes $v_1, \ldots, v_k$ to be added to the shortest path tree. We call the other nodes "free nodes".

We assume that for every $j \in [k]$, we have a search point $x^{(j)}$ encoding the shortest path $P_j$ to $v_j$. Further, we assume that for every $j \in [k]$ and every free vertex $v$ we have a search point encoding the path $P_j$ extended by attaching $v$ to $x^{(j)}j$.

Consider the search point $x^{(k)}$ lastly found. We call the nodes belonging to the path in $x^{(k)}$ "connected nodes". Note that a node cannot at the same time be free and connected, but there may be nodes that are neither free nor connected. We apply $3n \log n$ times the following unary unbiased operator. "Create the path $z$ from $x^{(k)}$ by pointing exactly one of the unconnected nodes to $v_k$. Let every other unconnected node point to itself." The operator is unbiased since every source-preserving automorphism of $x^{(k)}$ must fix the path in $x^{(k)}$, and hence must fix $v_k$.

With probability at least $1 - n^{-2}$ we have queried all $n - 1 - k$ possible attachments of free nodes to $v_k$. We compute the lowest cost for connecting one of them via $n_k$ and compare it with the lowest cost for connecting a free vertex via one of the vertices $v_1, \ldots, v_{k-1}$. Of all these connections, we choose the cheapest and store the corresponding search point.

Since we need to add $n - 1$ vertices, the search phase needs time $O(n^2 \log n)$.

For the construction phase, assume that we have learned the complete shortest path tree $T$. I.e., for every vertex $v$, we have a search point encoding the path in $T$ from the source to $v$. We want to construct $T$ explicitly. We start with the empty search tree (i.e., the search point where all nodes point to themselves), and add vertices in a depth-first manner with a particular order in which we traverse the children. To describe this order, we define for every node $v$ of $T$ the *lengths sequence* as follows. Consider the subtree $T_v$ of $T$ consisting of $v$ and all its descendants. The lengths sequence is the ordered sequence of the lengths of all paths from $v$ to a leaf in $T_v$, with the largest length coming first. We say that a node *has maximal length* if its length sequence is maximal with respect to lexicographic ordering.

Now we describe how to construct $T$ iteratively. We refer to the time between adding the $(k-1)$st and $k$-th node the *k-th phase*. At the end of each phase, we choose an *active node*, to which the next node is to be attached. We start with the source being active.

For the $(k+1)$st phase, assume we have already constructed a tree $T_k$ of size $k$, encoded in a search point $y^{(k)}$. We will make use of the unary operator $\texttt{attach}(y^{(k)})$, which creates a search point $z$ from $y^{(k)}$ by redirecting exactly one unconnected node to the active node. We will choose the active node in such a way that this operator is structure-preserving unbiased.

Consider all children of the active node in the shortest path tree $T$ that are not yet connected in $T_k$. Among those, choose a child $v$ of maximal length. Let $x_v$ be the search point storing the shortest path from $s$ to $v$. We add $v$ to $T_k$ as follows. Sample $z \leftarrow \texttt{attach}(y^{(k)})$. Skipping the details, we note that it is possible by binary unbiased operations to compute in constant time the number of edges in which $z$ and $x_v$ coincide. Hence we can decide whether the newly added node was $v$. We keep on sampling $z \leftarrow \texttt{attach}(y^{(k)})$ until we find a $z$ that adds $v$ to $y^{(k)}$, then setting $y^{(k+1)} \leftarrow z$. If $v$ has a child in $T$ that is not yet connected then we make $v$ the next active node. Otherwise, we ascend the tree from $v$ until we find a node that has such a child, and make this node active.

The expected time needed to add a new node is $O(n)$. Since $n$ nodes need to be added, the construction phase needs time $O(n^2)$.

It remains to be shown that the variation operator is structure-preserving unbiased for our choice of the active node. Assume that the algorithm has already added $k$ nodes forming a tree $T_k$, and the active node is $w$. We need to show that every permutation of the non-source nodes mapping the tree $T_k$ isomorphic to itself must map $w$ to itself. Assume that there is such an isomorphism mapping $w$ to a different node $w'$. Then $w'$ is contained in $T_k$, so it must be either a leaf, or it must have been active at some time. Let $u$ be the closest common ancestor of $w$ and $w'$, and let $v$ and $v'$ be the children of $u$ that are ancestor of $w$ and $w'$, respectively. Then the algorithm has explored $v'$ before $v$, so the length sequence of $v'$ was larger or equal to the length sequence of $v$. Moreover, the subtree with root $v'$ has been explored completely. But since $w$ is active, not all children of $w$ have been discovered yet, and so the subtree with root $v$ has been explored only partially. Together, the subtree of $v'$ in $T_k$ has strictly greater length sequence than the subtree of $v$ in $T_k$. Therefore, there is no isomorphism of $T_k$ mapping $v$ to $v'$. Consequently, there is no isomorphism of $T_k$ mapping $w$ to $w'$, because every isomorphism must preserve ancestor relations. This concludes the proof that the variation operator is structure-preserving unbiased.

Summing up, the algorithm needs time $O(n^2 \log n) + O(n^2) = O(n^2 \log n)$. □